\documentclass[default,iicol,pdflatex]{sn-jnl}

\usepackage{graphicx}
\usepackage{amsmath}
\usepackage{amssymb}
\usepackage{mathtools}
\usepackage{url}
\usepackage{xcolor}
\usepackage[algo2e,ruled]{algorithm2e}
\usepackage[caption=false,font=footnotesize]{subfig}
\usepackage{bbm}

\jyear{2021}%

\theoremstyle{thmstyleone}%
\theoremstyle{thmstyletwo}%
\newtheorem{remark}{Remark}%

\theoremstyle{thmstylethree}%

\begin{document}

\title[Article Title]{Urban Fire Station Location Planning using Predicted Demand and Service Quality Index}


\author*[1]{\fnm{Arnab} \sur{Dey}}\email{dey00011@umn.edu}

\author[2]{\fnm{Andrew} \sur{Heger}}\email{aheger@ci.victoria.mn.us}

\author[3]{\fnm{Darin} \sur{England}}\email{engl0124@umn.edu}

\affil*[1]{\orgdiv{Electrical and Computer Engineering}, \orgname{University of Minnesota, Twin Cities}, \orgaddress{\city{Minneapolis}, \postcode{55455}, \state{MN}, \country{USA}}}

\affil[2]{\orgdiv{Fire Chief}, \orgname{City of Victoria Fire Department}, \orgaddress{\city{Victoria}, \postcode{55386}, \state{MN}, \country{USA}}, \url{https://www.ci.victoria.mn.us/100/Fire-Department}}

\affil[3]{\orgdiv{Industrial and Systems Engineering}, \orgname{University of Minnesota, Twin Cities}, \orgaddress{\city{Minneapolis}, \postcode{55455}, \state{MN}, \country{USA}}, \url{https://cse.umn.edu/isye/darin-england}}


\abstract{In this article, we propose a systematic approach for fire
	station location planning. We develop machine learning
	models, based on Random Forest and Extreme Gradient Boosting, for demand prediction and
	utilize the models further to define a generalized index to
	measure quality of fire service in urban settings. Our model
	is built upon spatial data collected from multiple different
	sources. Efficacy of proper facility planning depends on
	choice of candidates where fire stations can be located
	along with existing stations, if any. Also, the travel time
	from these candidates to demand locations need to be taken
	care of to maintain fire safety standard. Here, we propose a
	travel time based clustering technique to identify suitable
	candidates. Finally, we develop an optimization problem to
	select best locations to install new fire stations. Our
	optimization problem is built upon maximum coverage problem,
	based on integer programming.  We further develop a two-stage stochastic optimization model to characterize the confidence in our decision outcome. We present a detailed
	experimental study of our proposed approach in collaboration
	with city of Victoria Fire Department, MN, USA. Our demand
	prediction model achieves true positive rate of 80\% and
	false positive rate of 20\% approximately. We aid Victoria
	Fire Department to select a location for a new fire station
	using our approach. We present detailed results on
	improvement statistics by locating a new facility, as
	suggested by our methodology, in the city of Victoria.}

\keywords{Facility planning, fire risk, extreme gradient boosting, optimization, random forest, risk
	prediction}



\maketitle

\section{Introduction}\label{intro}

Urban fires adversely affect the socio-economic growth and ecosystem health of any community. In $2019$, the National Fire Protection Association (NFPA) reported an estimated loss of \$$14.8$ billion and $3700$ civilian fire deaths in the U.S. \cite{ahrens2020loss}. According to the report, on average, a fire department responded to a fire incident every $24$ seconds and in total, local fire departments responded to an estimated $1.3$ million fires. Thus, building a safe community equipped with adequate fire safety measures is one of the most essential and challenging tasks of any fire department and local government.
However, it is very difficult to quantify the measure of safety, which in-turn results into complexity in determining if additional fire stations are required to be installed. This decision must be substantiated with all necessary spatio-temporal data, gathered from various different sources, such as accurate
details of historical emergency incidents from the fire department,
future patterns of land use from local government, parcel information
from geographic information system (GIS) (maintained by both US Census
Bureau and local authorities) and demographic information from US
Census Bureau. Combining all of this information together is a demanding task, due to differences in storage architectures across
different sources. For example, the address of a real property is
documented with different abbreviations and acronyms in different data
sources, which makes the task of joining the data sets very
difficult. Also, the accuracy of information on historical fire
incidents is poor and exacerbated by incompleteness. A few examples
include address mismatches, missing dates, and missing timestamps. Further, even if the requirement can be empirically justified using historical spatio-temporal data, determination of locations for new facilities is
a demanding task which must be strategically chosen based on
performance of the existing facilities as well as future community
growth and fire risk projections. Such prediction of future service demand is very
difficult due to its dependency on many different underlying
factors. Moreover, correlation between the predictive features makes the task
even more challenging. \textcolor{black}{We emphasize that quantifying public safety measures, and embedding it to devise a systematic approach to locate new fire stations is limited in existing literature. Moreover, the efficacy of such approaches from a data-mining perspective, including both confidential and publicly available dataset, is unexplored in existing works.}

To this end, we provide an approach that utilizes both
temporal and spatial data to select locations for future fire stations in
urban settings. Our methodology uses a Random
Forest \cite{breiman2001random} and Extreme Gradient Boosting \cite{chen2016xgboost} machine learning models and a
Density-based spatial clustering of applications with noise
(DBSCAN) algorithm \cite{ester1996density}.  For our
experimental case study, we partnered with City of Victoria
Fire Department (VFD), MN, USA. We apply our proposed
methodology to select the location of a future fire station
for VFD. Note that our approach is not specific to VFD and can
be used to analyze candidate locations for any urban area. The main
contributions of this article are summarized below:

(1) \textbf{Predictive demand model and service
	quality assessment}: We propose two machine learning (ML)
models, based on Random Forest (RF) \textcolor{black}{and Extreme Gradient Boosting (XGBoost)}, to predict future service
demand of urban areas from spatial data. \textcolor{black}{We present a detailed performance comparison between the models to choose the superior one.} The chosen ML model is
utilized to define a generalized service quality index ($SQI$)
that measures the quality of service provided by existing fire
stations at any demand point. \textcolor{black}{We propose to use the SQI as a new national fire service standard to assess the performance of any fire facility, which can be tailored to a specific locality.} We also present a travel time
based DBSCAN algorithm to identify candidate locations for new
facilities.  Our proposed methodology can be used to predict
and to analyze the performance of any fire department to
improve quality of service.

(2) \textbf{Optimization model for facility location
	selection}: We propose an optimization model, based on $SQI$
estimates, to select future fire stations. Our optimization
model accounts for spatial coverage, service prioritization
and service redundancy, thus presenting an end-to-end
strategic approach for fire station location selection. \textcolor{black}{Further, we propose a two-stage stochastic optimization model utilizing our demand prediction model and travel-time base DBSCAN algorithm. While the SQI-based optimization model establishes the utility of our proposed service quality index to be used as a new standardized index, the stochastic optimization model establishes the confidence in the decision outcome of our proposed models, thus providing a comprehensive overview of the merits of fire station location selection methodology.}

(3) \textbf{Impact on the city of Victoria Fire Department}: The work
is done in collaboration with Victoria Fire Department (VFD). VFD
utilizes the detailed performance analysis of the existing fire
station to properly allocate resources to improve the quality of
service. It can also help VFD to identify properties and locations
which require special attention in terms of fire safety measures. The
predictive models and the facility location study are used by VFD for
crafting the initial proposal of the new fire station
to be approved by the local authorities.
\section{Related Work}
\label{sec:1}
Systematic approaches for selecting sites for critical
infrastructure have been widely studied in the literature
\cite{turkoglu2019comparative}. In \cite{yao2019location}, a
bi-objective spatial optimization model is proposed to locate
an urban fire station by minimizing the number of stations to
be sited along with their distances from the service demand
clusters to ensure a required area coverage. A similar
approach is applied in \cite{aktacs2013optimizing} to prove
the efficacy of such algorithms in guiding local government
authorities to select a location for a new fire station.  A
multi-criteria decision making approach for planning of emergency
facilities, tuned with GIS
information, is proposed in \cite{nyimbili2021comparative}.
Site selection
based on spatial efficiency of existing and future facilities
is studied extensively in many prior works
\cite{church2016estimating,chevalier2012locating,murray2010advances}. A
GIS based approach for emergency facility location is the
primary focus of
\cite{soltani2019spatial,abdullahi2014spatial}. Various
optimization models to solve facility location problems, both
in the context of cost and service coverage optimizations, are
widely studied in the literature
\cite{erlenkotter1978dual,toregas1971location,abareshi2019bi}. In
\cite{bolouri2020minimizing}, an optimization methodology
based on response time of the fire stations is proposed to
select new facilities.

However, when there exists a constraint on the maximum number
of facilities that can be sited, determining the weights on
multiple options in decision-making can be challenging. Here,
prioritization of service demand locations based on future
risk prediction can be helpful \cite{o2017empirical}. Fire
risk prediction has been a primary focus of many researchers;
however, this work has mostly pertained to forest and wild
land fires
\cite{sevinc2020bayesian,eden2020empirical,choi2020fire,bui2018gis,salehi2016dynamic}. \cite{turco2018skilful}
presents an extensive study on the effect of climate on fire
prediction along with an analysis of the impact of different
seasons on burned area from global perspective. In
\cite{cheng2008integrated}, the authors propose a neural
network model to predict forest fires with spatio-temporal
data.

Urban fire risk prediction has received little attention
despite the critical nature of such
applications. \cite{singh2018dynamic} provides a data-driven
approach to predict commercial and residential building
fires. In \cite{madaio2016firebird}, a Random Forest algorithm
is used to generate predicted fire risk scores to aide the
local fire department in identifying properties that require
fire inspections. Such data-driven predictive models are
widely used in many prior works
\cite{agarwal2020big,jin2020ufsp,lau2015fire,salehi2018survey}. \cite{jin2020urban}
proposes an interesting approach to prediction of fire
incidents that is based on a deep neural network. Here, the
authors use spatio-temporal data from GIS to learn a
generative model. However, application of these models to
locate new facilities has not been extensively
studied. Moreover, other critical aspects of fire rescue
service, such as response time \cite{jaldell2017important},
need to be combined with prediction models. In summary, an unified approach, to select new fire station locations in urban settings, that considers fire service demand predictions and temporal variables with proper characterization of service priorities and redundancies is limited in the literature.

To this end, we propose an end-to-end fire station location
selection methodology that respects predicted demand as well
as travel time. We propose a Random Forest (RF) model to
predict demand for urban fire and emergency services demand
and combine it with travel time information to assess service
quality. We quantify the quality of service using our proposed
\textit{Service Quality Index} and further utilize it to
identify candidate locations for additional fire
stations. Finally, an optimization model is presented to
select the best option(s) among the candidate locations. In
summary, we devise a systematic approach for locating a new fire station
using GIS and temporal data.
Our demand prediction model is inspired by
\cite{madaio2016firebird,singh2018dynamic}; however, instead
of restricting the model to residential and commercial
buildings, we consider all types of properties present in a
typical city. We apply our proposed methodology to aid VFD
with site selection for the new fire station in the city of
Victoria. Historical fire incident data is provided by VFD
which is augmented with GIS and other demographic data
collected from multiple different sources. We substantiate the
efficacy of our method with extensive experimental results.

The rest of the paper is organized as follows: The proposed service quality index and the demand prediction models are explained in Section~\ref{sec:serv_quality_assessment}. Section~\ref{sec:candidate_determination} describes the algorithm to choose suitable candidate locations for new station installation. Two optimization problem formulations, to select new fire station locations from the candidates, are introduced in Section~\ref{sec:location_selection}. Efficacy of our proposed approach is illustrated with a detailed case study in Section~\ref{sec:case_study}. Section~\ref{sec:conclusions} presents the concluding remarks.
\section{Service Quality Assessment}\label{sec:serv_quality_assessment}
The service quality of existing fire stations is an important
indicator that influences the choice of candidate locations for future
facilities. However, in the context of urban fire safety, there is no
standard approach to assess the service quality that considers
demographic factors such as age of residents, income, property types,
population density, etc. Apart from demographic factors, another
important aspect to measure fire safety is travel time from the
responding station to the incident location. A longer travel time
can result in significant damage or even loss of life. In this
context, we propose a generalized service quality index for each
property within the service area of a fire station, considering both
demographic factors and travel time, as described in the following
sections.
\subsection{Service Quality Index}\label{subsec:sqi_def}
Let $\mathcal{J}$ denote the set of demand locations (properties) and
let $\mathcal{I}$ denote the set of fire stations within the city
boundary. We define the Service Quality Index ($SQI$) of a demand location
$j\in \mathcal{J}$ with respect to a fire station $i\in \mathcal{I}$
as follows:
\begin{align}\label{eq:sqi_per_station}
SQI_{ji} \coloneqq P(j)\hat{T}(j,i),
\end{align}
where $P(j) \in [0,1]$ denotes the probability of demand request at a
location $j$ and $\hat{T}(j,i)$ denotes the normalized travel
time from fire station $i$ to the demand location $j$. The normalized
travel time can be calculated as follows:
\begin{align}\label{eq:travel_time_norm}
\hat{T}(j,i) &= \frac{T_{actual}(j,i)}{T_{norm}},
\end{align}
where $T_{norm}$ is a normalization factor to ensure $\hat{T}(j,i) \in [0,1]$ and $T_{actual}(j,i)$ is
the actual travel time from fire station $i$ to demand location $j$
which can be calculated using GIS. A lower value of $SQI_{ji}$
indicates better service. Note that, $SQI_{ji}$ depends on the
probability of demand request $P(j)$ of property location $j$, which
usually depends on factors like local population density,
property type (residential, business etc.), property age, and
demographic factors such as residents' age, median income,
etc. Moreover, $SQI_{ji}$ also considers the response time for emergency
personnel to reach a location, a crucial factor in quality of service
\cite{national1720nfpa}.
\begin{remark}In the trivial case of no existing fire station,
$\hat{T}(j,i)$ is undefined. Here, one can take $SQI_{ji}$ values to
be equal to $P(j)$ and the rest of the analysis follows.
\end{remark}
\begin{remark}The trade-off between low demand probability at a
location with a long travel time from $i$ and high demand probability
at a location with a short travel time from $i$ ensures a balance
between area coverage and emphasis on demand request. However, the
interpretation of better (or worse) service quality can be tuned per
the facility planners' requirements as we describe next.
\end{remark}

Availability of multiple fire stations ideally enhances the service
quality at a location. However, it is very difficult to characterize,
quantify, and predict the availability status of all the stations at
the time of demand request. Also note that the probability of a demand
request usually depends on demographic and property characteristics,
and hence is independent of the location of fire stations. Therefore, we
factor out the dependency on a particular fire station by taking the
minimum of $SQI_{ji}$ values over all fire stations
$i \in \mathcal{I}$. Therefore, the Service Quality Index of a
property $j$ can be written as
\begin{align}\label{eq:sqi}
SQI(j) = \min_{i \in \mathcal{I}}SQI_{ji}.
\end{align}

The interpretation of better/worse service quality can be done by
categorizing the demand locations into low, medium, and high quality
locations based on their $SQI$ values. Let the sets
$\mathcal{J}_l, \mathcal{J}_m, \mathcal{J}_h \subseteq \mathcal{J}$
denote the set of indices of the demand locations with low, medium and
high quality of service which are defined as follows:
\begin{align}\label{eq:sqi_category}
&j \in \mathcal{J}_h, \ \text{ if }0 \leq SQI(j) < \tau_l,\nonumber\\
&j \in \mathcal{J}_m, \ \text{ if }\tau_l \leq SQI(j) < \tau_h,\\
&j \in \mathcal{J}_l, \ \text{ if }\tau_h \leq SQI(j) \leq 1,\nonumber
\end{align}
where $\tau_l, \tau_h$ are user-defined constants in $\mathbb{R}$ with $0<\tau_l<\tau_h \leq 1$.
Note that, the choice of $\tau_l$ and $\tau_h$ provides
an interpretation of fire service quality at a location based on
planning constraints. For example, one can calculate the values of
$\tau_l,\tau_h$ based on maximum allowable travel time (which depends
on the type of fire station \cite{national1720nfpa}) and an acceptable demand
request probability that is to be considered as high demand. Thus, our
objective is to identify candidate locations to install additional
fire stations such that the $SQI$ values of $j\in \mathcal{J}_l$ are
improved.
\subsection{Demand Prediction Model}
Calculation of $SQI$ requires a fire service demand prediction model. In addition to the ability to accurately predict
the probability of a service request for a future emergency event, it
is also important to gain insights about the predictors and their
effects on demand for services.  Such information can be utilized for
resource planning and staff management. In this context, tree based
models are suitable as they can be used to quantitatively assess the
importance and the effect of the predictors. We propose to use Random
Forest (RF) \cite{breiman2001random} \textcolor{black}{and Extreme Gradient Boosting (XGBoost) \cite{chen2016xgboost} algorithms, along with a detailed performance comparison}, to predict the
probability of a demand request $P(j)$ for each property
$j \mspace{-2mu}\in\mspace{-2mu} \mathcal{J}$. In building the \textcolor{black}{prediction} model\textcolor{black}{s}, we consider various
demographic factors and property features as described in
Table~\ref{tbl:ml_feat_risk}. \textcolor{black}{Note that, service demand prediction, with an objective of fire station location selection, is a static analysis. Unlike short-term demand prediction models based on online learning algorithms \cite{liu2020ambulance} in a dynamic environment, our model requires long-term demand prediction. Thus we include only spatial factors in our prediction models. The temporal factors, governing the dynamic nature of service demand, are analyzed in \cite{dey2020planning} to aide VFD in resource planning and staff management.}
The properties included in our analysis are categorized into four
categories based on their usage as shown in
Table~\ref{tbl:prop_type}. For our case study, the types of land use are
obtained from \cite{victoria2019}.
\begin{table}[!h]
	\begin{center}
	\caption{Categorization of properties}\label{tbl:prop_type}
	\begin{tabular}{ll} 
		\toprule
		Category		& Included property types\\[0.5ex] 
		\midrule
		Residential		& $1-3$ units residential properties,\\
		& apartments with $4+$ units\\
		Commercial		& Warehouses, retail outlets,\\
		& factories etc.\\
		Institution     & Churches, schools, Govt. offices,\\
		& police stations etc.\\
		Park			& Parks, lakes, wetlands, agricultural\\
		& lands, golf courses, vacant lands etc.\\
		\botrule
	\end{tabular}
	\end{center}
\end{table}

The predictor and response variables of our \textcolor{black}{prediction} model are described in Table~\ref{tbl:ml_feat_risk}. 
\begin{table}[!h]
	\begin{center}
	\caption{Demand prediction models feature description (Variable type notation: P $\equiv$Predictor, R $\equiv$Response)}\label{tbl:ml_feat_risk}
	\begin{tabular}{lcll} 
		\toprule
		Variable name & Type      & Range                 & Description\\[0.5ex] 
		\midrule
		Land value 	    & P 	& $\mathbb{R}^+$    & Estimated land\\
		&		&							&value ($\times$\$$10,000$).\\
		Land size 	        & P 	& $\mathbb{R}^+$  & Land size (acres).\\
		Num units             & P  & $\mathbb{N}$  & Number of units\\
		&		&								&in a property.\\
		Prop. Age            & P  & $\mathbb{N}$  & Age of property\\
		&					&					&at time of incident.\\
		Resi. Age	& P & $\mathbb{R}^+$	& Residents' Median\\
		&&&age in the block\\
		&			&					& where the property\\
		&&&is located.\\
		Population	& P	& $\mathbb{R}^+$	& Average population\\
		&&&of the block where\\
		&			&								& property is located.\\
		Prop. Type 	& P & $0-3$    & $0 \equiv$ residential,\\
		&&(factor)&$1 \equiv$ commercial\\
		&			&								& $2 \equiv$ institution,\\
		&&&$3 \equiv$ park\\
		\midrule
		Demand   & R  & $0-1$    & $1$ if any\\
		&&(binary)&incident occurred,\\
		&&&$0$ otherwise\\
		\botrule
	\end{tabular}
	\end{center}
\end{table}

\begin{remark}The age of a property is calculated from the year it is built and
the year when an incident is reported. For example, if a property
built in $2010$ reports a fire event in $2015$, then
`Age'$=5$. Multiple events at the same location at different points in
time are treated separately. For the properties with no reported
incidents, the age is calculated with reference to the year of writing
the article (year $2021$).
\end{remark}

Once the \textcolor{black}{prediction} model\textcolor{black}{s are} built, we obtain the demand request
probabilities $P(j)$ using the class probability estimates of the
samples corresponding to the positive labeled class (class of samples
with response variables labeled as $1$; Refer to
Table~\ref{tbl:ml_feat_risk}). We present the detailed performance
measures of our proposed \textcolor{black}{prediction} models and variable importance in
Section~\ref{subsec:result_rf} for our case study.

\noindent Upon finding the demand request probabilities, it remains to compute
travel time from a fire station to a property in order to completely
characterize $SQI$ values for each property. We calculate travel time
using the Open Street Map-Based Routing Service (OSRM) package \cite{osrm}
in R \cite{R} with traffic flow disabled to account for the relaxed
traffic rules for fire service vehicles. Once the $SQI$ values are
calculated, we need to identify suitable geographic locations for
additional fire stations. In the next section, we discuss our proposed
methodology to select such candidate locations.
\section{Determining candidate
	locations}\label{sec:candidate_determination}
The probability of a demand request depends on demographic and
property characteristics and is independent of the location of a fire station.
Thus, from~(\ref{eq:sqi_per_station}), we note that
improvement of $SQI$ values can be achieved by placing additional fire
stations in close proximity to areas with properties having high $SQI$
values (poor service) with respect to the existing fire stations. Let
$\mathcal{I}_c$ denote the index set of candidate locations for new fire stations.
We aim to find out the geographic locations of candidate
fire stations $i_c \in \mathcal{I}_c$ such that
$SQI(j),~j\in \mathcal{J}_l$, is minimized (service becomes
better). In this context, our objective is to locate high density
clusters of properties in $\mathcal{J}_l$. Once the clusters are
found, we can choose to place the new fire stations within the
clusters by considering other qualitative features such as road
accessibility, terrain condition, availability of vacant space,
etc. Note that in order to improve the fire service, finding the
clusters that have high densities of properties $j \in \mathcal{J}_l$,
should be based on the travel time rather than euclidean distance
(which is not suitable for measuring distance between points on a
sphere) or haversine distance. For example, even if two properties are
close to each other in terms of their longitude and latitude, the time
to travel from one to another may be large due to various
factors such as road inaccessibility or traffic congestion.
Moreover, the existence of outliers (for example, a few properties
far from the main city), and arbitrarily shaped clusters are
well-known problems in spatial clustering that cannot be handled by
centroid-based clustering algorithms such as K-Means
\cite{bishop2006pattern}. Another challenge associated with clustering
algorithms such as K-Means is the required specification of the number of
clusters. Although there exist various methods to empirically find the
optimal number of clusters \cite{alpaydin2020introduction}, they are
not always reliable in spatial clustering
\cite{ester1996density}. Therefore, we propose to use the
\textit{Density-based spatial clustering of applications with noise}
(DBSCAN) algorithm \cite{ester1996density} with distance measure
replaced with travel time. We group properties based on the
travel time between pairs of properties. Let $n$ denote the total
number of properties in $\mathcal{J}_l$. We first compute the travel
time between each pair of properties in $\mathcal{J}_l$ using the OSRM
package \cite{osrm} in R \cite{R}. The travel time information is
stored in a symmetric matrix
$\boldsymbol{T} \in \mathbb{R}_+^{n \times n}$, where each element
$\boldsymbol{T}[i,j]$ denotes the actual travel time from $i$ to $j$,
$T_{actual}(i,j)$. The longitude and latitude values of each property
are stored in $\boldsymbol{P} \mspace{-2mu}\in\mspace{-2mu} \mathbb{R}^{n \times 2}$ such that
the $i^{th}$ row of $\boldsymbol{P}$ corresponds to the $i^{th}$
row of $\boldsymbol{T}$. Our travel time based DBSCAN requires two
parameters, $\epsilon$ and $\delta$. $\epsilon$ denotes the
$\epsilon$-neighborhood of a property $j$, which is defined as the set
$\{k \in \mathcal{J}_l\mid \boldsymbol{T}[k,j] \leq \epsilon\}$. $\delta$
denotes the minimum number of properties required to be present in the
$\epsilon$-neighborhood of $j$ to be considered a part of a
cluster. Note that in travel time based DBSCAN, the number of clusters
is found during runtime and is not required to be specified a priori.
Once the clusters are obtained, we choose to
locate the candidate fire stations close to the centroids of the
clusters (for our case study, we account for the factors such as
availability of vacant space based on the future city development plan
\cite{victoria2019}, road accessibility, terrain, etc.) The travel time based DBSCAN algorithm is shown in
Algorithm~\ref{alg:tt_dbscan}. Next, we present the optimization
model to select the best location(s) among the candidate sites.

\begin{algorithm2e}[h]
	\SetKwFunction{getResources}{getResources}
	\SetKwFunction{FP}{FindCandidate}
	\SetKwFunction{BP}{BP}
	\SetKwProg{Fn}{Function}{:}{}
	\SetKwInOut{Input}{Input}{}
	\SetKwBlock{Initialize}{Initialize:}{}
	\SetKwBlock{STEPONE}{STEP 1:}{}
	\SetKwBlock{STEPTWO}{STEP 2:}{}
	\SetKwBlock{Repeat}{Repeat for $ k = 1,2, \dots$}{}
	\SetKw{Continue}{continue}
	\newcommand{\inparallel}{\textbf{In parallel}}
	\Fn{\FP($\boldsymbol{P}\mspace{-3mu}\in\mspace{-3mu} \mathbb{R}^{n\mspace{-2mu}\times\mspace{-2mu} 2},\boldsymbol{T} \mspace{-2mu}\in\mspace{-2mu} \mathbb{R}^{n\mspace{-2mu}\times\mspace{-2mu} n}$,$\epsilon$,$\delta$)}{
		$cluster\_id \leftarrow 0$; $outlier \leftarrow -1$;\\
		\lFor{$i=1$ \KwTo $n$}{$\boldsymbol{L_n}[i] \leftarrow undefined$}
		\For{$i = 1$ \KwTo $n$}{
			\lIf{$\boldsymbol{L_n}[i] \neq undefined$}{\Continue}
			$\boldsymbol{N} \leftarrow \{j:\boldsymbol{T}[i,j] \leq \epsilon\}$\\
			\If{$\mid\boldsymbol{N}\mid < \delta$}{
				$\boldsymbol{L_n}[i] \leftarrow outlier$;\\
				\Continue;
			}
			$cluster\_id \leftarrow cluster\_id + 1$;\\
			$\boldsymbol{L_n}[i] \leftarrow cluster\_id$;\\
			$\boldsymbol{S_n} \leftarrow \{s: s \in \boldsymbol{N}, s \neq i\}$;\\
			\ForEach{$j \in \boldsymbol{S_n}$}{
				\If{$\boldsymbol{L_n}[j] = outlier$}{
					$\boldsymbol{L_n}[j] \leftarrow cluster\_id$;
				}
				\lIf{$\boldsymbol{L_n}[j] \neq undefined$}{\Continue}
				$\boldsymbol{N} \leftarrow \{k:\boldsymbol{T}[j,k] \leq \epsilon\}$;\\
				$\boldsymbol{L_n}[j] \leftarrow cluster\_id$;\\
				\lIf{$\mid\boldsymbol{N}\mid < \delta$}{\Continue}
				$\boldsymbol{S_n} \leftarrow \boldsymbol{S_n} \cup \boldsymbol{N}$;
			}
		}
		\For{$c=1$ \KwTo $cluster\_id$}{
			$\boldsymbol{C_f}[c] \leftarrow centroid(\{\boldsymbol{P}[k,:]:\boldsymbol{L_n}[k] = c\})$;
		}
	}
	\Return{$\boldsymbol{C_f}$};
	\caption{Finding Candidate Locations}
	\label{alg:tt_dbscan}
\end{algorithm2e}
\section{Fire Station Location Selection}\label{sec:location_selection}
Installation of new fire stations along with existing facilities can
cause the service areas of the fire stations to overlap. It is
desirable to locate new fire stations in such a way that
the overlap of service areas is minimized and the aggregated
coverage area is maximized. In this context we
first introduce the concept of catchment area and then formulate an
optimization problem utilizing the $SQI$ values and the catchment
areas of each candidate location, to select best location(s) among the candidates to install new fire station(s). \textcolor{black}{Further, in a facility planning model, involving demand probabilities, it is required to characterize the stochasticity of decision outcome. Therefore, we propose a two-stage stochastic optimization formulation, based on probabilistic reward function, to analyze the stochastic performance of the our model. Note that, both of our proposed optimization formulation can be utilized to select the best location(s) to install new fire station(s) while the two-stage stochastic optimization model provide more insight into the confidence in the decision.}
\subsection{Catchment Area}\label{subsec:catchment_area}
\begin{figure}[!h]
	\centering
	\includegraphics[scale=0.34,trim={0.2cm 0.1cm 0.1cm 0.1cm},clip]{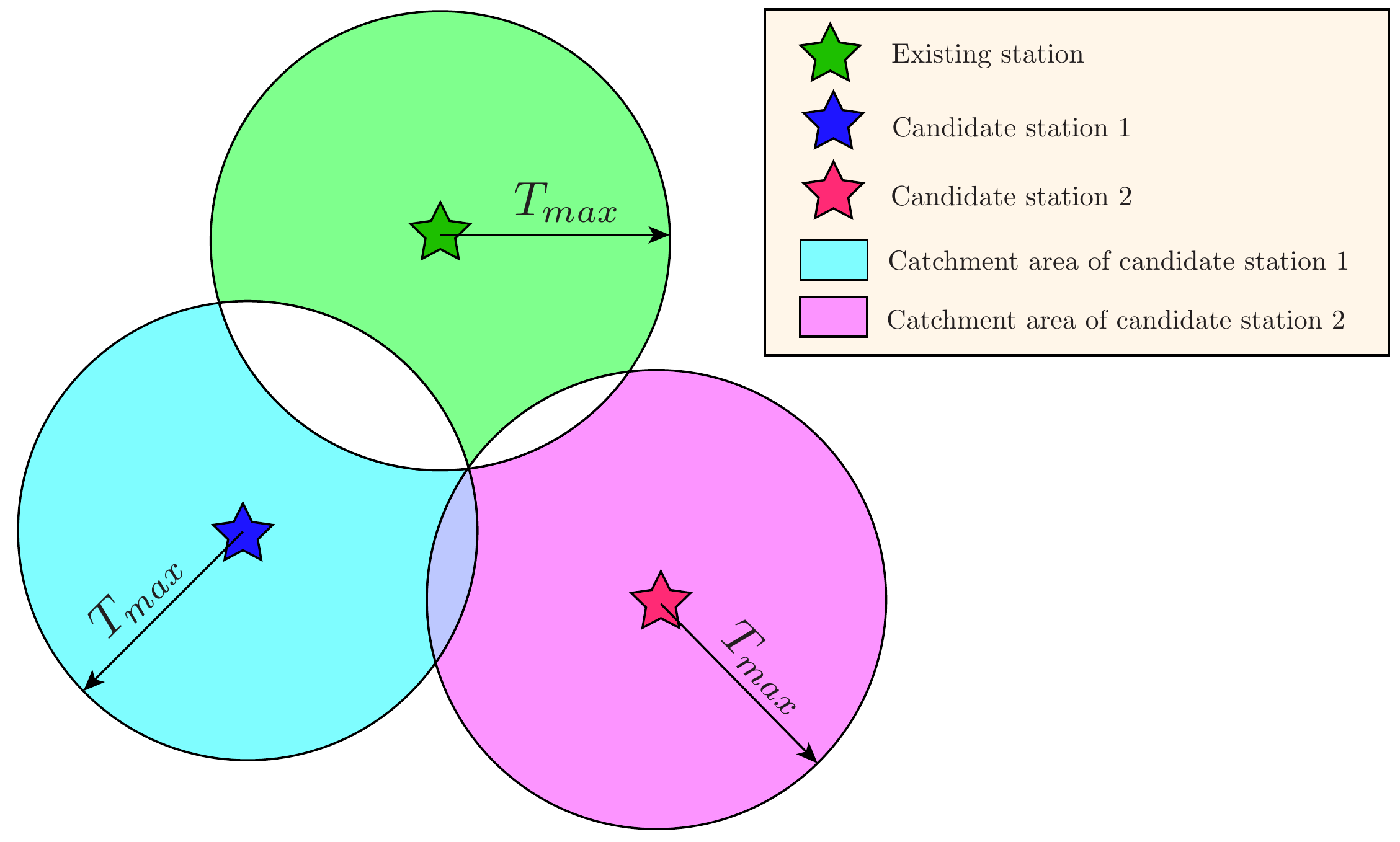}
	\caption{Catchment area with one existing and two candidate stations. Circles with $T_{max}$ radius denote areas that can be reached from respective stations within $T_{max}$ time}
	\label{fig:catchment_example}
\end{figure}

The fire service in a city usually maintains a performance bound on
the time required to travel from the station to the incident
location. The bound depends on the type of fire station
\cite{national1720nfpa} and the local government. Let $T_{max}$ denote
the specified bound on the travel time. With this
constraint, the candidate locations of new fire stations $i_c \in \mathcal{I}_c$
should allow the number of demand points $j \in \mathcal{J}$ for which
$T_{actual}(j,i_c) \leq T_{max}$ to be maximized. Essentially, the catchment area of a
candidate location $i_c$ includes properties $j \in \mathcal{J}$ that
cannot be accessed from any existing station $i \in \mathcal{I}$
within $T_{max}$ but can be accessed from $i_c$ within the specified
travel time limit of $T_{max}$. An illustrative example of catchment
areas with one existing fire station and two candidate fire stations
is shown in Fig.~\ref{fig:catchment_example}.
In this example, the stations are located at the center of circles
covering areas that can be reached within $T_{max}$ units of time from
the center. Note that the catchment areas of the candidate stations
are the areas that do not overlap with the existing station. To ensure
good area coverage along with $SQI$ improvement, we propose to place
the future stations within the cluster of high $SQI$ properties that
cover the maximum number of properties within their respective
catchment areas. In the next section we present the optimization
model.  It is derived from the maximum area coverage problem
\cite{church1974maximal}.
\subsection{SQI Based Optimization Problem Formulation}\label{subsec:opt_formulation}
The best locations for future fire stations are those that maximize
the number of properties with high $SQI$ (poor service by the existing
fire stations) within their respective catchment areas. A property
with higher $SQI$ (poorer service by existing fire stations) is given
more priority relative to a property with lower $SQI$ (better service
by existing fire stations). Therefore, we propose to weigh the
properties according to their $SQI$ values. Let us assume that it is
required to locate at most $p$ number of additional fire stations. For
each candidate location $i \in \mathcal{I}_c$ we consider a binary
decision variable $x_i$. It takes a value of $1$ if candidate
location $i$ is selected and $0$ otherwise. For each
property $j \in \mathcal{J}$ we consider another binary decision
variable $y_j$ which takes a value of $1$ if that property is
within the catchment area of any candidate location
$i \in \mathcal{I}_c$, and $0$ otherwise. For each property $j$, $N_j$
denotes the set of candidate stations whose catchment areas cover that
property. In other words, the property $j$ can be reached within
maximum allowable travel time from any of the candidate locations in $N_j$. Our
objective is to obtain values for $x_i$ and $y_j$ that maximize the weighted sum
$\sum_{j \in \mathcal{J}} SQI(j)y_j$. Hence, our proposed weighted
maximum coverage optimization problem can be stated as follows:
\begin{align}
\text{maximize} & \sum_{j \in \mathcal{J}}SQI(j)y_j\\
\text{subject to} & \sum_{i \in N_j} x_i \geq y_j,\ \text{ for all }j \in \mathcal{J},\\
& \sum_{i \in N} x_i \leq p,\\
& x_i \in \{0,1\}\ \text{ for all }i \in \mathcal{I}_c,\\
& y_j \in \{0,1\}\ \text{ for all }j \in \mathcal{J},
\end{align}
where
\begin{align*}
\mathcal{J} &= \text{ set of all demand locations within the city},\\
\mathcal{I}_c &= \text{Set of candidate fire stations},\\
p &= \text{Number of fire stations to be located},\\
\hat{T}(j,i) &= \text{Normalized travel time from fire station $i$},\\
& \hspace{0.5cm}\text{to demand location $j$},\\
\hat{T}_{max} &= \text{Normalized maximum allowable travel time},\\
N_j &= \{i \in \mathcal{I}_c \mid \hat{T}(j,i) \leq \hat{T}_{max}\},\\
x_i &= \begin{cases}
1\ \text{ if }i \in N_j, \text{ for all }j\in \mathcal{J},\\
0\ \text{ otherwise,}
\end{cases}\\
y_j &= \begin{cases}
1\ \text{ if } \hat{T}(j,i) \leq \hat{T}_{max}, i \in \mathcal{I}_c,\\
0\ \text {otherwise.}
\end{cases}
\end{align*}
Let
$\boldsymbol{x}^* \coloneqq \{x^*_i : x^*_i \in \{0,1\}, i \in
\mathcal{I}_c\}$ and
$\boldsymbol{y}^* \coloneqq \{y^*_j : y^*_j \in \{0,1\}, j \in
\mathcal{J}\}$ denote the optimal values of the decision variables. The
selected locations for the new fire stations are those
for which $x^*_i = 1$.
\subsection{Two-stage Stochastic Optimization Problem Formulation}\label{subsec:stoch_opt_formulation}
\textcolor{black}{Our proposed two-stage stochastic optimization problem is an iterative process involving a stochastic reward function $R_t(i)$ associated with each candidate location $i \mspace{-4mu}\in\mspace{-4mu} \mathcal{I}_c$ in iteration $t \mspace{-4mu}\in\mspace{-4mu} \mathbb{N}$. We associate a random variable $X_{j,t}$ with each demand location $j \in \mathcal{J}$ in $t^{th}$ iteration. We assume that, for all $t \in \mathbb{N}$ and $j \in \mathcal{J}$, the set of random variables $\{X_{j,t}\}$ are independent and identically distributed according to Bernoulli distribution with $\text{Pr}(X_{j,t}=1) = P(j) = 1-\text{Pr}(X_{j,t}=0)=1-(1-P(j))$. Note that, such a definition assigns a value of $1$ or $0$ to $X_{j,t}$ based on the demand request probability, $P(j)$. Let $\mathcal{C}_i \coloneqq \{j \in \mathcal{J} \mid \hat{T}(j,i) \leq \hat{T}_{max}, i \in \mathcal{I}_c\}$ denote the catchment area of candidate $i \in \mathcal{I}_c$. We define the reward function as
\begin{align}\label{eq:reward_func}
	R_t(i) \coloneqq \sum_{j \in \mathcal{C}_i} X_{j,t}.
\end{align}
Our proposed approach is as follows: at iteration $t \geq 1$, in the first stage, the candidate, $i_t \in \mathcal{I}_c$, having maximum expected reward till $(t-1)^{th}$iteration, $Q_{t-1}(i_t)$, is chosen. We compute the expected reward, till $t^{th}$ iteration, corresponding to each candidate location as follows:
\begin{align}\label{eq:expected_reward}
	Q_t(i) \coloneqq \frac{\sum_{k=1}^{t} \mathbbm{1}_{(i_k = i)}R_k(i)}{\sum_{k=1}^{t}\mathbbm{1}_{(i_k=i)}},
\end{align}
where $\mathbbm{1}_{(i_k=i)} = 1$ if candidate $i$ is chosen at iteration $k$, and $0$ otherwise. Thus,
\begin{align}\label{eq:first_stage}
	\textstyle i_t = \textstyle \text{argmax}_{i \in \mathcal{I}_c} Q_{t-1}(i).
\end{align}
In the next stage, all demand locations in it's catchment area, $\mathcal{C}_{i_t}$, are triggered and the values $\{X_{j,t} \mid j \in \mathcal{C}_{i_t}\}$ are probabilistically determined and the reward $R_t(i_t)$ is evaluated as per~(\ref{eq:reward_func}). Once $R_t(i_t)$ is evaluated, we update the expected reward for all $i \in \mathcal{I}_c$ as per~(\ref{eq:expected_reward}).}

\noindent\textcolor{black}{We employ an epsilon-greedy approach to enable exploration of all candidates to find the optimal candidate that maximizes the reward, that is, with probability $(1-\epsilon)$ we choose $i_t$ as per~(\ref{eq:first_stage}) and with probability $\epsilon$, we randomly choose a candidate from $\mathcal{I}_c$. Upon termination of our proposed algorithm, based on a maximum iteration count $t_{max}$, the candidate having maximum expected reward is chosen as the optimal location to install a new fire station. Note that, if $p$ number of fire stations are to be located, where $p > 1$, the first $p$ candidates corresponding to expected rewards sorted in descending order, can be chosen. The sets $\mathcal{I}_c$ and $\mathcal{C}_i$ are derived as described in Section~\ref{sec:candidate_determination} and Section~\ref{subsec:catchment_area} respectively and $\epsilon$ is tuned empirically. The entire algorithm is summarized in Algorithm~\ref{alg:opt_cand}.
\begin{algorithm2e}[h]
	\SetKwFunction{GOC}{getOptimalCand}
	\SetKwProg{Fn}{Function}{:}{}
	\SetKwInOut{Input}{Input}{}
	\SetKwBlock{Initialize}{Initialize:}{}
	\SetKwBlock{STEPONE}{STEP 1:}{}
	\SetKwBlock{STEPTWO}{STEP 2:}{}
	\SetKwBlock{Repeat}{Repeat for $ k = 1,2, \dots$}{}
	\SetKw{Continue}{continue}
	\newcommand{\inparallel}{\textbf{In parallel}}
	\Fn{\GOC($\{P(j),j\in \mathcal{J}\}, \epsilon, \mathcal{I}_c$, $\{\mathcal{C}_i, i \in \mathcal{I}_c\},p$)}{
		\For{$i=1$ \KwTo $\mid \mathcal{I}_c\mid$}{
		Initialize $R_0(i), Q_0(i)$;\\
		$\boldsymbol{Q}[i] \leftarrow Q_0(i)$;
		}	
		\For{$t = 1$ \KwTo $t_{max}$}{
			$i_t \leftarrow \begin{cases}\text{argmax}_{i \in \mathcal{I}_c} Q_{t-1}(i), \text{ with Pr}(1-\epsilon)\\\text{any }i \in \mathcal{I}_c \text{ with Pr}(\epsilon)\end{cases}$;\\
			\For{$j=1$ \KwTo $\mid C_{i_t}\mid$}{
				$X_{j,t} \sim \text{Bernoulli}(P(j))$;
			}
			Update $R_t(i_t)$ as per~(\ref{eq:reward_func});\\
			Update $Q_t(i_t)$ as per~(\ref{eq:expected_reward});\\
			$\boldsymbol{Q}[i_t] \leftarrow Q_t(i_t)$;
		}
	$\boldsymbol{I}[\cdot] \leftarrow \text{argsort}_i\boldsymbol{Q}[i]$;
	}
	\Return{$\{\boldsymbol{I}[1],\boldsymbol{I}[2],\ldots,\boldsymbol{I}[p]\}$};
	\caption{Finding Optimal Candidate}
	\label{alg:opt_cand}
\end{algorithm2e}
We now present a detailed case study of our
entire fire station location methodology described in the previous
sections.}
\section{Case Study: City of Victoria, MN}\label{sec:case_study}
To apply our proposed methodology,
we partnered with the city of Victoria fire Department (VFD) to select a suitable
location for a new fire station. The city
of Victoria is located in Carver county of Minnesota, USA. The city
enjoys a solid economic and agricultural foundation. This has
catalyzed rapid urbanization and population growth in the past few
years. The increased acceptance of remote-work situations will further
add to the city's growth. Per the US Census Bureau \cite{uscensus},
the population of the city rose from $7400$ in $2010$ to $9170$ in
$2017$. The current population (at the time of writing) is
approximately $10000$ and this number is expected to increase by
$68$\% by the year 2040 \cite{victoria2019}. Currently, Victoria is
being served by one fire station. The expected growth and economic
development has caused concern among the city's leadership that the
current infrastructure may be insufficient to provide adequate public
safety services, including those services provided by the fire
department. Therefore, VFD initiated a resilient community project
(RCP) in $2019$ to assess the performance of the existing facility and
to identify a location for an additional fire station. The new
facility will enhance (not replace) the capabilities of the existing
fire station. A data exploratory analysis, as a part of the RCP
initiative, is performed in \cite{rcp9a} to find spatio-temporal
demand statistics. Following the statistics and the city guidelines,
as presented in \cite{rcp9a}, we apply our location selection
methodology to provide a comprehensive guidance on future station
location. Our database features and the detailed description of our
results are presented in the next sections.
\subsection{Description of Data}\label{sec:data_prep}
The entire database used in this study is comprised of data from
multiple different sources. Data containing the details (date,
locations, response time etc.) of historical incidents that occurred
between January $2010$ and October $2020$ are provided by VFD. \textcolor{black}{In particular, the dataset contains: (1) incident types (fire, explosion and overheat, Rescue and emergency medical service, Hazardous condition, service call, good intent call, false alarm, natural disaster, and special intent type), (2) address of incident location, (3) date and time, (4) Response time, (5) fire service vehicle used. As there are reported incidents for these properties between $2010$ and $2020$, these properties are designated as demand requesting locations.} We use Carver county's open data portal \cite{carver2020parcel} to further augment this dataset with properties with no reported incidents, \textcolor{black}{and these properties are designated as no demand requesting locations}. The entire dataset contains $5334$ samples of which $47.9$\% have reported historical incidents. Parcel
information for each property location is also obtained from \cite{carver2020parcel}. The parcel information includes address, land size (acres), estimated land value
(\$), property usage category (see Table~\ref{tbl:prop_type}), and
year of construction. This dataset is augmented with block-wise
average population and median age of residents obtained from the US
census bureau and Federal Communications Commission \cite{censusblock,fcc2020}. For the spatial analysis, the latitude and longitude of
the property locations are obtained from the US Census Bureau Geocoding
Service \cite{census2020geocode}. However, some locations cannot be
geocoded due to incorrect address entries, such as zip code and
address line mismatch, which are primarily caused by human
errors. These entries are matched by address lines in the Carver County
Active911 database \cite{carver2020active911} to obtain the geographic
locations. The US census bureau geocoder along with the Active911
database reduce the number of missing data due to common errors.
Every record is matched by property address to create the fully
processed database. There are several sources of missing
data, mainly due to
misspellings and abbreviations of the addresses (such as `highway' is
termed as `hwy' / `hgwy' / `highway' and sometimes misspelled as
`hgihway', `drive' is represented as `Dr'/`drv'/`drive' etc.). Some of
the common errors are taken care of in the data cleaning step; however,
several different types of errors leads to some amount of missing
data. ($4.2$\% of the entire dataset). We performed our proposed station location selection analysis on
this dataset and the results are presented in the subsequent sections.
\subsection{Demand Prediction using Random Forest and XGBoost}\label{subsec:result_rf}
\begin{figure}[!h]
	\centering
	\captionsetup[subfigure]{labelformat=empty}
	\subfloat[]{\includegraphics[scale=0.28,trim={1.5cm 3.6cm 2.5cm 0cm},clip]{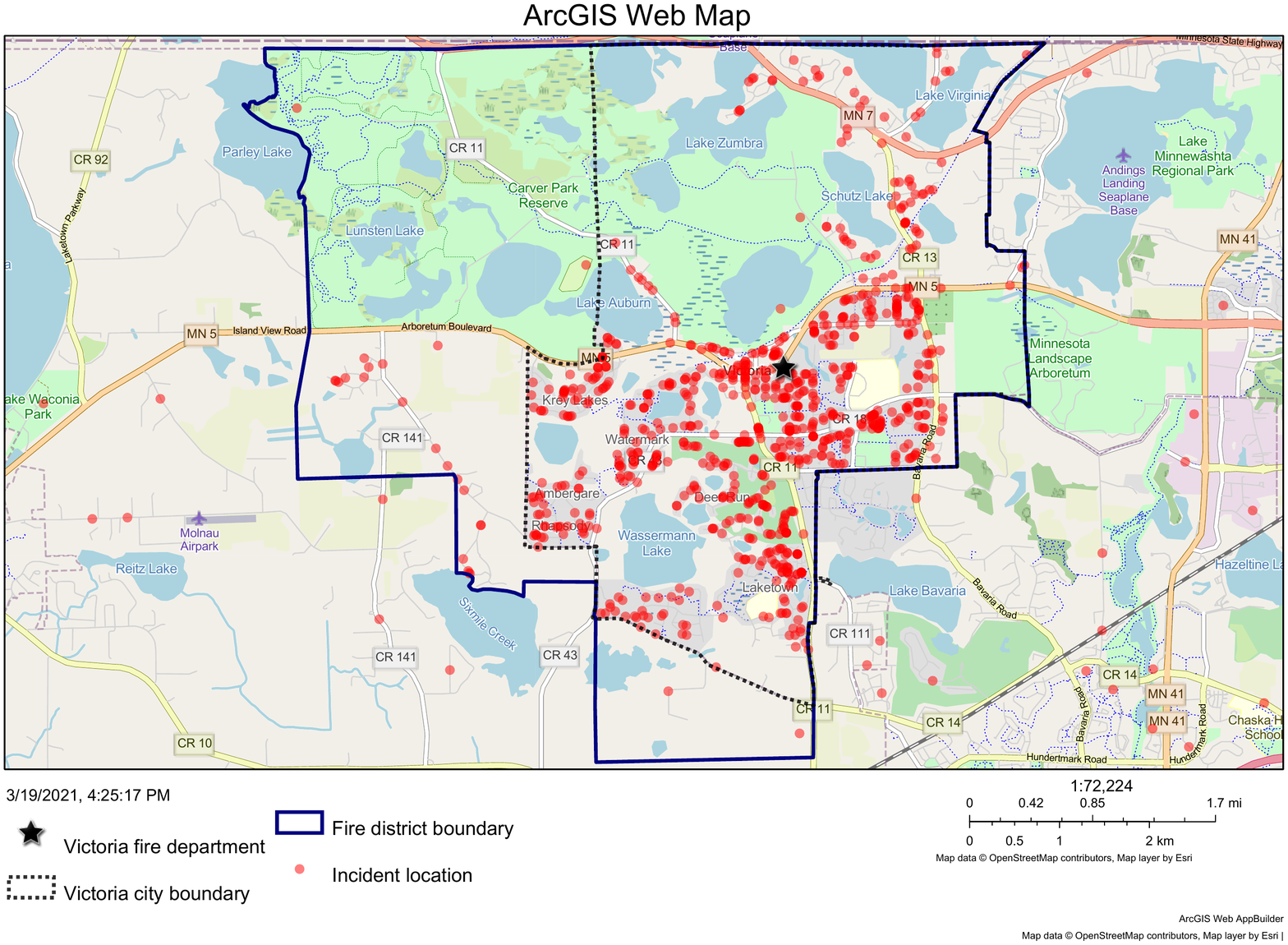}}\hspace{0.0cm}\vspace{-0.8cm}
	\subfloat[]{\includegraphics[scale=0.6,trim={1.0cm 0.5cm 10cm 18cm},clip]{map_incidents.pdf}}
	\caption{Locations of all incidents occurred during $2010-2020$ (till Oct): Incidents are concentrated in southern region and in residential areas. Existing facility is marked with black star icon and the locations are marked with red dots. Blue solid line indicates fire district boundary. Black dashed line indicates Victoria city boundary}
	\label{fig:map_incidents}
\end{figure}

Fig.~\ref{fig:map_incidents} shows the locations of all incidents
that occurred in $2010-2020$ (until Oct 2020) on an open street map. At the
time of writing, the City of Victoria has dense residential areas in
the southeast \cite{victoria2019}. The concentration of the incident
locations in the southeast, as can be seen in
Fig.~\ref{fig:map_incidents}, indicates that the service demand is
mostly concentrated in the residential areas. All of the maps are
created using ArcGIS \cite{arcgis2020}. The shapefiles of the fire
district and city boundary are taken from
\cite{fireshapefile,township}. The incidents outside the fire district
are typically serviced in cooperation with other fire departments,
while some of the events that are outside the city boundary, but
inside the Victoria fire district, are serviced by the VFD.
For rest of our analysis, we consider only those properties that are within the VFD fire district and served by VFD only, in case of emergency events. We combine the properties with history of reported incidents and other properties with no historical event. The entire dataset, with the sample features shown in Table~\ref{tbl:ml_feat_risk}, is used to build our proposed prediction models for demand request prediction. 

\noindent We implement our RF model using the Scikit-learn package
\cite{scikit-learn} and XGBoost model using the xgboost package in Python. $80$\% of the total number of samples
are used for training the model and the remaining 20\% are used for
performance testing and fine tuning.  Table~\ref{tbl:ml_grid_risk} \textcolor{black}{and~\ref{tbl:ml_grid_risk_xgb}}
shows the set of parameters used for grid search, with $10$-fold cross
validation, to find the best set of parameters of RF and XGBoost model respectively. \textcolor{black}{For our RF model, we set minimum number of samples required to split a node of a tree to $2$, and enable bootstrapping from the entire training sample set while building the trees.} The selected
parameters in the right-most column resulted in the best accuracy.
The performance of the prediction models are presented in
Table~\ref{tbl:ml_perf_risk}.
\begin{table}[!h]
	\begin{center}
	\caption{Grid parameters for Demand prediction RF model training}\label{tbl:ml_grid_risk}
	\begin{tabular}{rcc} 
		\toprule
		Parameters   & Grid search & Selected\\
		& parameters & parameter\\[0.5ex] 
		\midrule
		Criterion 	    & Entropy, Gini index   & Gini index\\
		Max tree depth 	        & $2, 4, 6, 8, 10$  & $8$\\
		Number of trees             & $100, 200, 300, 400$  & $300$\\
		No. of features            & $\sqrt{\text{No. of total features}}$   & $3$\\
		randomly chosen	&&\\
		Min samples in& $30, 40, 50$ & $30$\\
		leaf nodes &&\\
		\botrule
	\end{tabular}
	\end{center}
\end{table}
\begin{table}[!h]
	\begin{center}
		\caption{Grid parameters for Demand prediction XGBoost model training}\label{tbl:ml_grid_risk_xgb}
		\begin{tabular}{rcc} 
			\toprule
			Parameters   & Grid search & Selected\\
			& parameters & parameter\\[0.5ex] 
			\midrule
			Learning rate & $0.01, 0.1, 0.2, 0.3$ & $0.2$\\
			Max tree depth 	        & $2, 4, 6, 8, 10$  & $4$\\
			Number of trees             & $100, 200, 300, 400$  & $100$\\
			Gamma            & $0,1,2,3,4$   & $3$\\
			Alpha & $0.001, 0.002, 0.003$ & $0.002$\\
			Lambda & $1,2,3$ & $1$\\
			Subsample & $0.7, 0.8, 0.9$ & $0.8$\\
			Random feature & $0.5, 0.6, 0.7$   & $0.6$\\
			ratio per tree&&\\
			\botrule
		\end{tabular}
	\end{center}
\end{table}
\begin{table}[!h]
	\begin{center}
	\caption{Demand prediction model performance measures}\label{tbl:ml_perf_risk}
	\begin{tabular}{rcc} 
		\toprule
		Performance measure   & RF & XGBoost\\[0.5ex] 
		\midrule
		Training accuracy 	    & $74$\% & $87$\%\\
		Test accuracy 	        & $70$\% & $78$\%\\
		Out-of-Bag score             & $69$\% & NA\\
		True positive rate: Training set            & $69$\% & $89$\%\\
		False positive rate: Training set   & $22$\% & $11$\%\\
		True positive rate: Test set    & $65$\% & $80$\%\\
		False positive rate: Test set   & $26$\% & $21$\%\\
		AUC: Training set   & $0.83$ & $0.94$\\
		AUC: Test set   & $0.79$ & $0.85$\\
		\botrule
	\end{tabular}
	\end{center}
\end{table}
\begin{figure}[!h]
	\centering
	\subfloat[]{\includegraphics[scale=0.9,trim={0.32cm 0cm 0.1cm 0cm},clip]{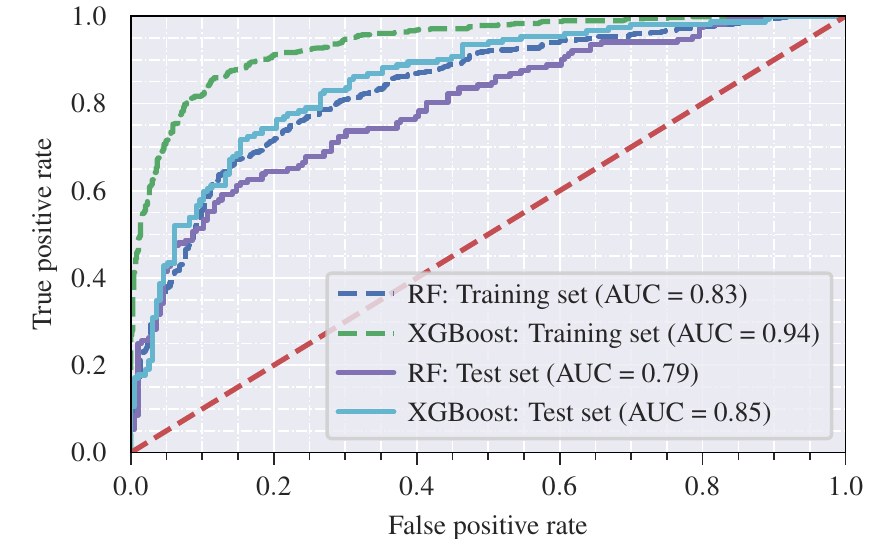}\label{fig:ROC_risk}}\hspace{0.2cm}
	\subfloat[]{\includegraphics[scale=0.9,trim={0.3cm 0cm 0.1cm 0cm},clip]{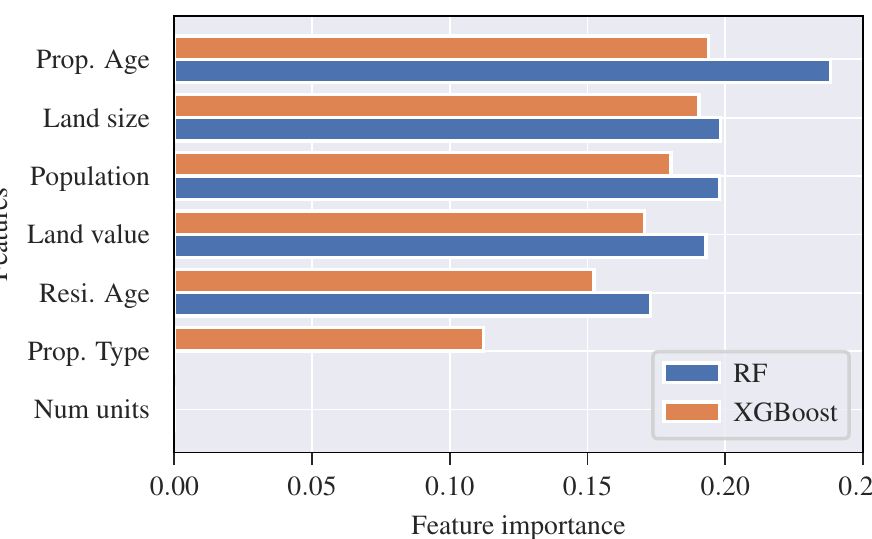}\label{fig:feat_imp_risk}}
	\caption{Demand Prediction ML model performance: (a) ROC curve. AUC values are $\geq 0.7$, which indicates the efficacy of our proposed model. (b) Feature importance shows that property age and residents age are the two most important factors governing the demand request}
	\label{fig:ROC_risk_feat_imp_risk}
\end{figure}

\begin{figure}[!h]
	\centering
	\captionsetup[subfigure]{labelformat=empty}
	\subfloat[]{\includegraphics[scale=0.28,trim={1.05cm 3.6cm 1.84cm 1.25cm},clip]{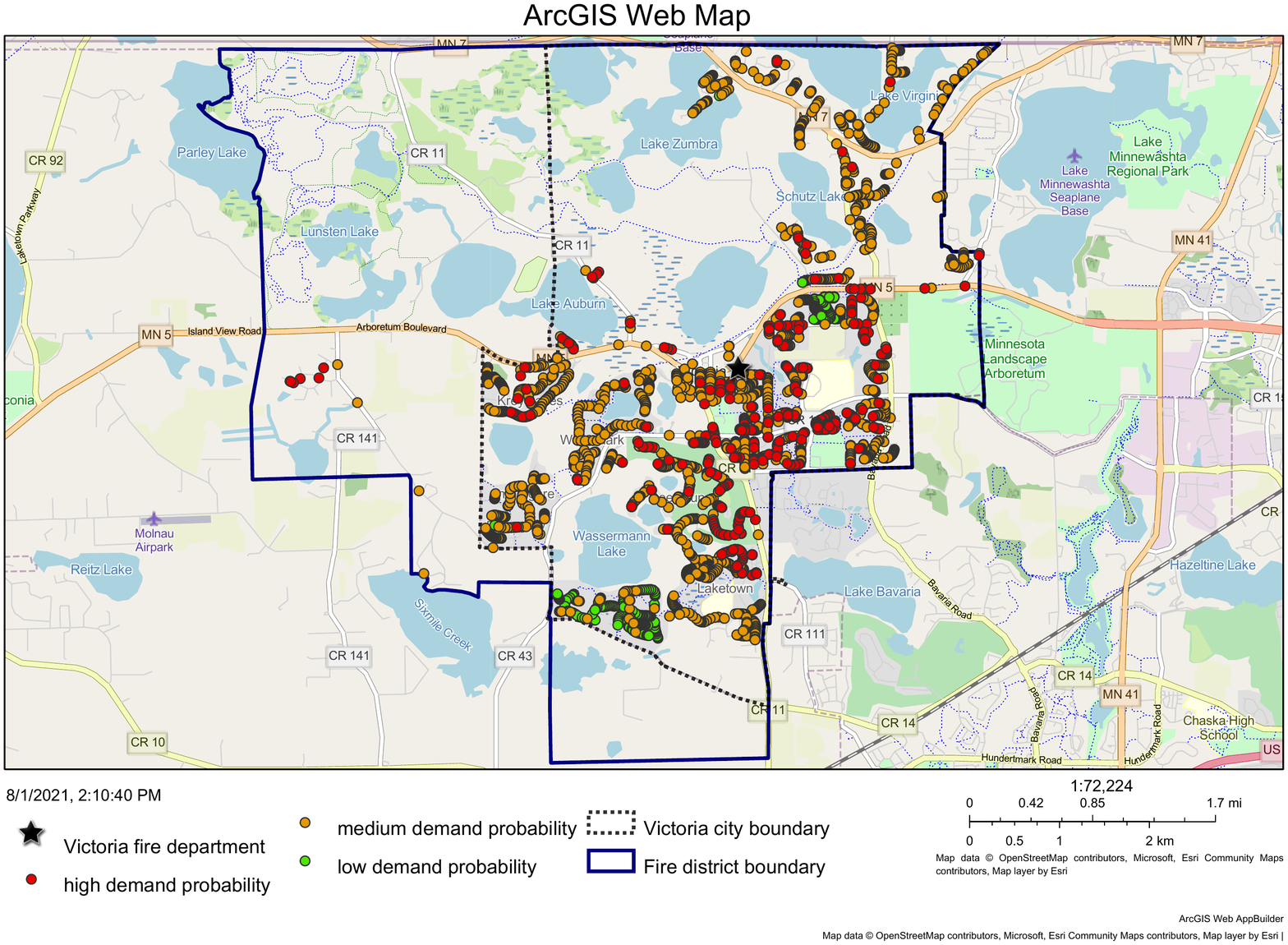}}\hspace{0cm}\vspace{-0.8cm}
	\subfloat[]{\includegraphics[scale=0.42,trim={1.05cm 0.5cm 10cm 18cm},clip]{map_demand_prob.pdf}}
	\caption{Spatial property distribution based on probability of demand request: Existing facility is marked with black star icon and low, medium, and high service quality properties are marked in red, orange, and green dots respectively. Blue solid line indicates fire district boundary. Black dashed line indicates Victoria city boundary}
	\label{fig:map_demand_prob}
\end{figure}

The ROC curve \cite{james2013introduction} is shown in
Fig.~\ref{fig:ROC_risk}. It can be seen that \textcolor{black}{both of} our models perform very
well in predicting the demand request of different property types. \textcolor{black}{Further, our extreme gradient boosting model performs significantly better than the RF model in terms of accuracy and true and false positive rates. RF model exhibits a true positive rate (TPR) of $69$\% and $65$\% for the training and the test set
respectively while with XGBoost they are improved to $89$\% and $80$\% respectively. The false positive rate (FPR) values of $11$\% and $21$\%, of XGBoost model, are relatively
low and significantly better than those of the RF model.} While TPR is associated with performance guarantees of our $SQI$
based \textcolor{black}{stochastic optimization based }planning algorithms, FPR is closely related to the cost
effectiveness of VFD and local government, as it can result in
unnecessary investments in safety measures. \textcolor{black}{We derive impurity based feature importance, as
illustrated in Fig.~\ref{fig:feat_imp_risk}, to explain the
effect of demographic and property features on the service demand
request}. Here, we observe that the most important features in
determining service demand are property age, median age of residents
in the block where the property is located, land size, population
at the block level, and estimated land value. We note that this additional
information can be utilized by city of Victoria to further analyze
future service demand with projected growth
\cite{victoria2019}. \textcolor{black}{As our XGBoost model performs better than the RF model, }we use the class probability estimates of our XGBoost
model as the demand request probabilities for the calculation of
$SQI$. The raw class probabilities are scaled to have values between
$0$ to $1$ using min-max normalization. We categorize the demand
request probabilities into three categories as shown in
Table~\ref{tbl:prob_percentage}, in order to tune $\tau_l$ and
$\tau_h$ that are required
for~(\ref{eq:sqi_category}). Table~\ref{tbl:prob_percentage} shows the
number of properties falling under each category as a percentage of
total number of properties within the VFD fire district. The spatial
distribution of the same is presented in
Fig.~\ref{fig:map_demand_prob}.
\begin{table}[!h]
	\begin{center}
		\caption{Demand Request Probability of Properties}\label{tbl:prob_percentage}
		\begin{tabular}{lc} 
			\toprule
			Demand probability   & Number of properties\\[0.5ex] 
			\midrule
			Low ($0 \leq Pr < 0.35$) 	    & $18.1$\%\\
			Medium ($0.35 \leq Pr < 0.65$) 	        & $61.9$\%\\
			High ($0.65 \leq Pr \leq 1$)             & $20$\%\\
			\botrule
		\end{tabular}
	\end{center}
\end{table}
After obtaining the demand request probabilities, the travel time from
VFD to each property are required in order to compute the $SQI$
values \textcolor{black}{and the catchment areas}. In the next sections, we present the comprehensive results of
our new fire station location selection methodology based on $SQI$
values \textcolor{black}{and two-stage stochastic optimization model} as described in Section~\ref{sec:candidate_determination} and
Section~\ref{sec:location_selection}.
\subsection{Fire Station Location Selection}
\begin{figure}[!h]
	\centering
	\captionsetup[subfigure]{labelformat=empty}
	\subfloat[]{\includegraphics[scale=0.28,trim={1.05cm 3.6cm 1.84cm 1.25cm},clip]{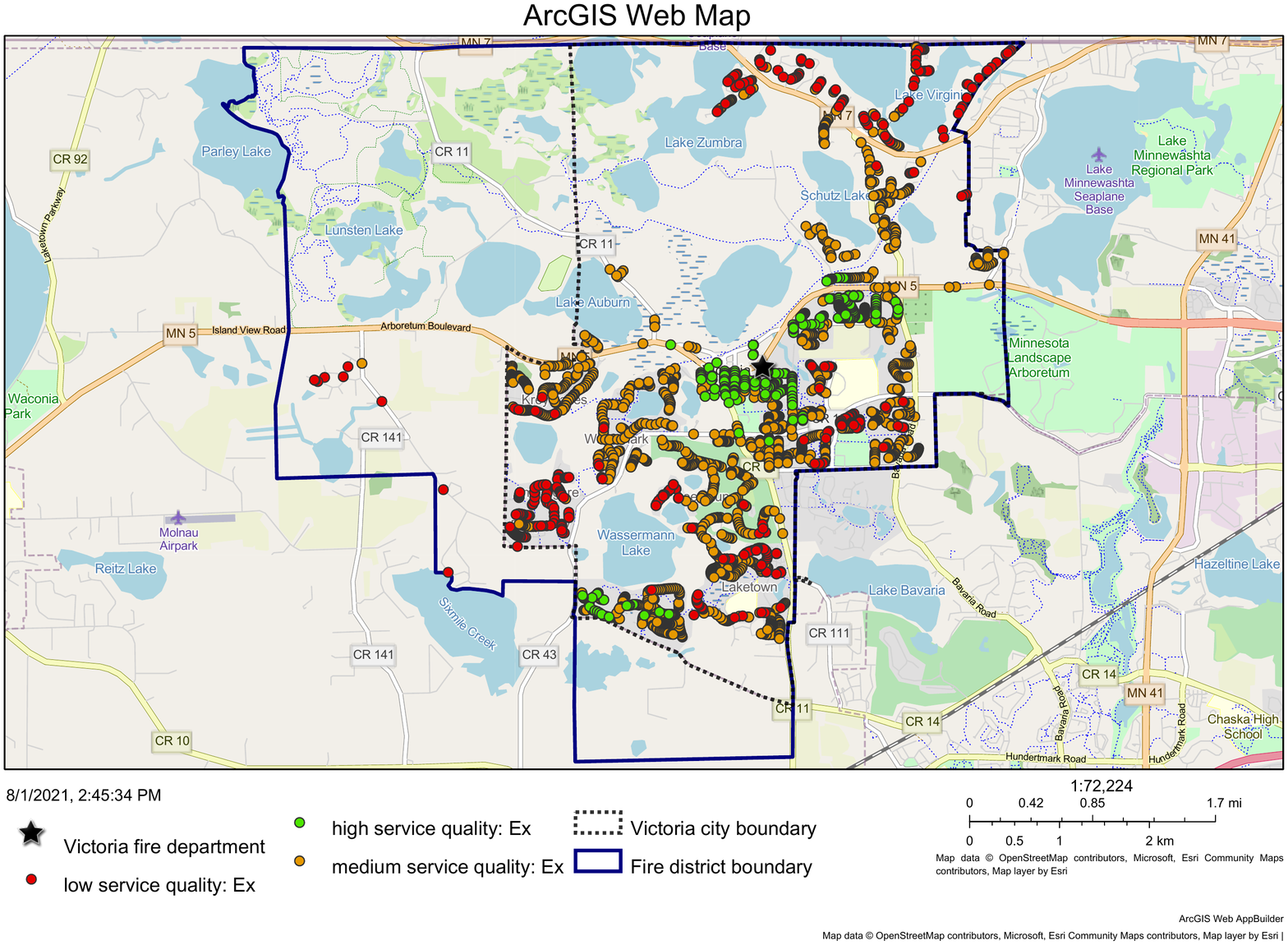}}\hspace{0cm}\vspace{-0.8cm}
	\subfloat[]{\includegraphics[scale=0.42,trim={1.05cm 0.5cm 10cm 18cm},clip]{map_service_quality.pdf}}
	\caption{Service Quality with existing fire station: Existing facility is marked with black star icon and low, medium, and high service quality properties are marked in red, orange, and green dots respectively. Blue solid line indicates fire district boundary. Black dashed line indicates Victoria city boundary}
	\label{fig:map_service_quality}
\end{figure}

Once the probability of service demand of all properties are
predicted using the prediction model and the travel time from VFD to each property
is obtained from OSRM \cite{osrm}, we calculate the $SQI$ values of
the properties using~(\ref{eq:sqi}). VFD has a performance goal that
travel time is not to exceed $4$ minutes. Therefore, we have taken
$T_{max} = 4$. The other parameters that are required for
Algorithm~\ref{alg:tt_dbscan} and the for the SQI-based optimization model, described in Section~\ref{subsec:opt_formulation}, are listed
in Table~\ref{tbl:loc_params}. Note that the parameters
$\tau_l,\tau_h$ are derived from the demand request probability
categorization shown in Table~\ref{tbl:prob_percentage}
and Equation~(\ref{eq:sqi_per_station}).
\begin{table}[!h]
	\begin{center}
	\caption{Location selection parameter values}\label{tbl:loc_params}
	\begin{tabular}{rcc} 
		\toprule
		Parameter   & Values & Units\\[0.5ex] 
		\midrule
		$T_{max}$ 	    & $4$ & minutes\\
		$T_{norm}$ 	        & $20$ & minutes\\
		$\tau_l$             & $0.05$ & None\\
		$\tau_h$            & $0.16$ & None\\
		$\epsilon$   & $2$ & minutes\\
		$\delta$    & $80$ & None\\
		$p$			& $1$ & None\\
		\botrule
	\end{tabular}
	\end{center}
\end{table}
We calculate the $SQI$ values and categorize the properties into low,
medium, and high service quality properties as
per~(\ref{eq:sqi_category}). The total number of properties falling under
each category as a percentage of total number of properties within
the fire district is presented in Table~\ref{tbl:sqi_percentage_ex}.
\begin{table}[!h]
	\begin{center}
		\caption{$SQI$ percentages}\label{tbl:sqi_percentage_ex}
		\begin{tabular}{lc} 
			\toprule
			Quality of Service   & Number of properties\\[0.5ex] 
			\midrule
			Low ($\tau_h \leq SQI \leq 1$) 	    & $12$\%\\
			Medium ($\tau_l \leq SQI < \tau_h$) 	        & $72.9$\%\\
			High ($0 \leq SQI < \tau_l$)             & $15.1$\%\\
			\botrule
		\end{tabular}
	\end{center}
\end{table}
The geographical distribution of properties is shown in
Fig.~\ref{fig:map_service_quality}. Here, a visual comparison between
Fig.~\ref{fig:map_demand_prob} and Fig.~\ref{fig:map_service_quality}
is useful to provide an interpretation of the remark made in
Section~\ref{subsec:sqi_def}. It is observed that most of the
properties with medium demand request probabilities, but far from
the existing fire station, are marked as receiving low quality of service
while the properties with high demand but close to the existing fire
station are marked as receiving better service.

\begin{figure}[!h]
	\centering
	\captionsetup[subfigure]{labelformat=empty}
	\subfloat[]{\includegraphics[scale=0.28,trim={1.05cm 3.6cm 1.84cm 1.25cm},clip]{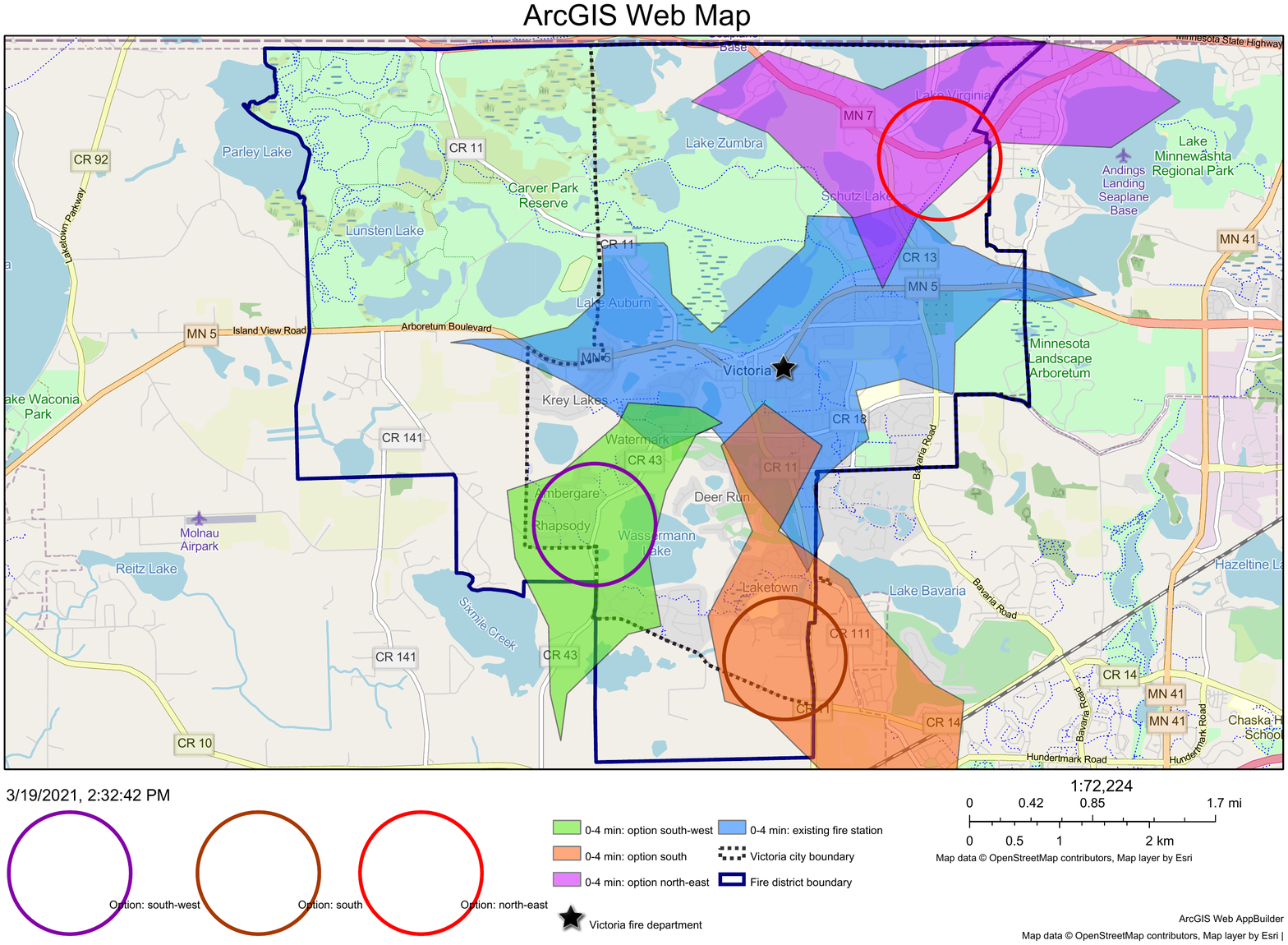}}\hspace{0cm}\vspace{-0.8cm}
	\subfloat[]{\includegraphics[scale=0.4,trim={1.05cm 0.2cm 10cm 18cm},clip]{map_overlap_combined.pdf}}
	\caption{Catchment areas of candidate locations based on $4$-minute drive time polygon: Overlap is least for SW thus indicating least redundancy in service along with the existing facility. Existing facility is marked with black star icon and the candidate locations are shown with large circles to indicate broader zones. Blue solid line indicates fire district and black dashed line indicates Victoria city boundary}
	\label{fig:map_overlap_combined}
\end{figure}

\begin{figure}[!h]
	\centering
	\subfloat[]{\includegraphics[scale=0.9,trim={0.18cm 0cm 0.2cm 0.07cm},clip]{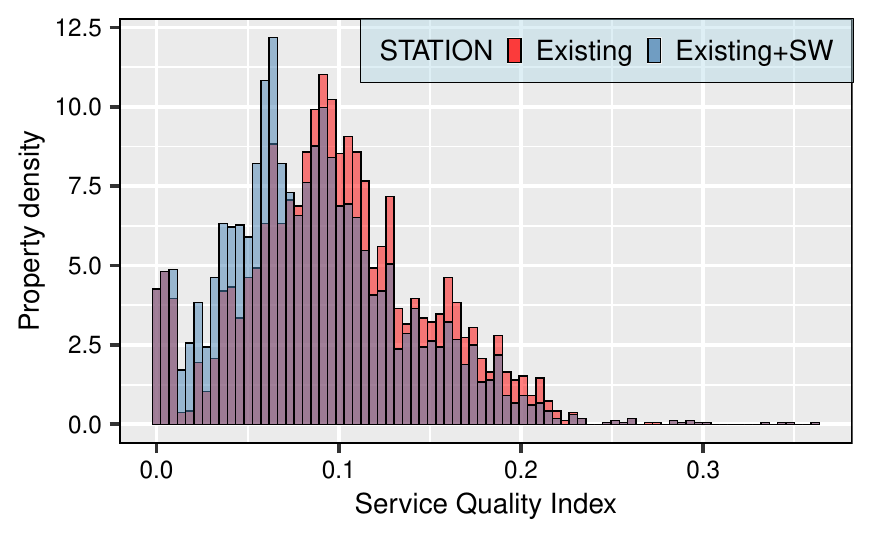}\label{fig:sqi_hist_ex_sw}}\hspace{0.1cm}
	\subfloat[]{\includegraphics[scale=0.9,trim={0.18cm 0cm 0.2cm 0.07cm},clip]{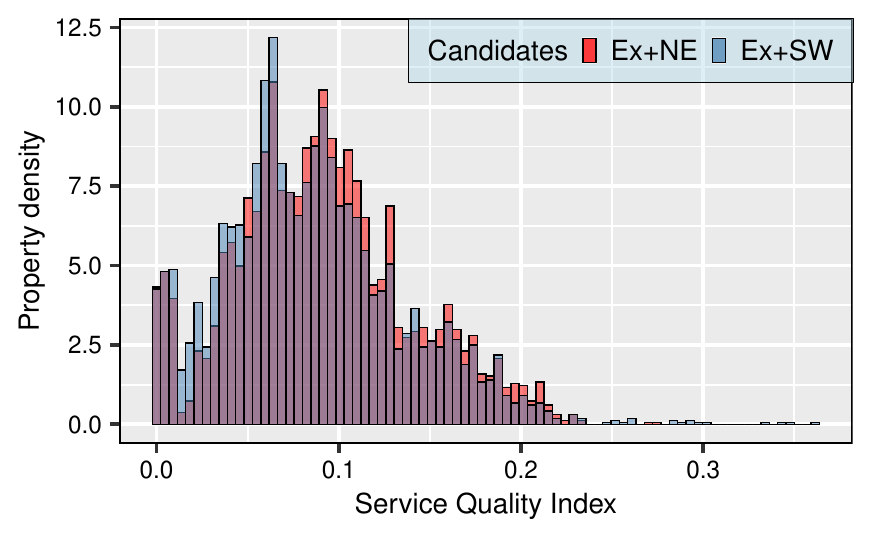}\label{fig:sqi_hist_sw_ne}}\hspace{0.1cm}
	\subfloat[]{\includegraphics[scale=0.9,trim={0.18cm 0cm 0.2cm 0.07cm},clip]{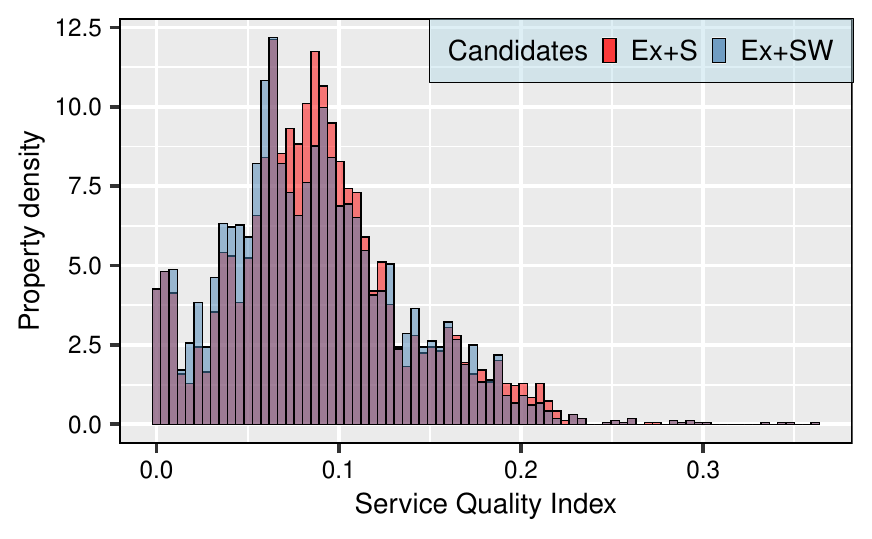}\label{fig:sqi_hist_sw_s}}
	\caption{(a) SW candidate along with existing fire station (marked in blue) increases the density towards the head of the distribution, thus improving number of properties with better service quality, compared to that with only existing station (marked in red). (b) SW candidate along with existing station (marked in blue) improves number of properties with better service quality compared to NE candidate with the existing station (marked in red). (c) Similar improvement can be seen by comparing the histograms corresponding to SW with existing station (marked in blue) and S with existing station (marked in red)}
	\label{fig:sqi_hist}
\end{figure}

Now, we apply Algorithm~\ref{alg:tt_dbscan} on the properties marked
with red in Fig.~\ref{fig:map_service_quality} to find the candidate
locations. Algorithm~\ref{alg:tt_dbscan} is implemented using
the Scikit-learn package \cite{scikit-learn} in Python. The result is
shown in Fig.~\ref{fig:map_overlap_combined} along with the $4$-minute
drive time polygons of the candidates. We only show the approximate
locations of the candidates with circles on the map to conform to
the confidentiality of actual locations as requested by VFD. Based on
Algorithm~\ref{alg:tt_dbscan}, we find three candidate locations in the fire
district: (a) northeast, (b) southwest and (c) south. For the rest of our
analysis, we denote the candidates by:
\begin{enumerate}[1.]
	\item Northeast (NE)
	\item Southwest (SW)
	\item South (S).
\end{enumerate}
Note that the non-overlapped $4$-minute travel time polygons between
SW, S, and NE with the existing station form the catchment areas of
the candidates. We aim to apply our SQI-based optimization
methodology, as described in Section~\ref{subsec:opt_formulation},
with respect to these catchment areas. Properties within the catchment
areas of SW, S, and NE are obtained using `sf' package \cite{r_sf} in
R. VFD seeks a location for one additional fire station, i.e. $p=1$. 
We remark that although the project with VFD requires 
\textit{one} additional station, our methodology can be applied
without modification to more general settings for any value of
$p$. The optimization problem is solved using ILOG CPLEX Optimization
Studio V20.1 \cite{studio2020v20}. The solution to the optimization
problem indicates that SW candidate is the one that maximizes the
number of properties with high $SQI$ (lower service quality, marked
red in Fig.~\ref{fig:map_service_quality}) within its catchment
area. A detailed comparison of the number of properties in each $SQI$
category as a percentage of total number of properties is presented in
Table~\ref{tbl:sqi_percentage_comparison}. The existing fire station
is abbreviated as `Ex' and the expected performance, in case of
an additional fire station with the existing one, is tabulated. Note
that while the existing station provides high quality service to $15.1$\%
of properties, an additional fire station at SW can
enhance it to $23.8$\%. Thus, we can expect an improvement of $53$\%
over the existing performance. However, the installation of an additional
station at S or NE location results in an approximate $23-33$\% improvement
over the existing one. Similarly, in the case of low service quality
properties, SW location provides an improvement of approximately $25$\%,
while S and NE both provide an approximate $20$\% improvement.
\begin{table}[!h]
	\begin{center}
		\caption{$SQI$ percentages}\label{tbl:sqi_percentage_comparison}
		\begin{tabular}{lcccc} 
			\toprule
			Service Quality   & Ex & Ex+SW & Ex+S & Ex+NE\\[0.5ex] 
			\midrule
			Low 	   &$12$\% & $9$\% & $9.4$\% & $9.6$\%\\
			Medium 	        &$72.9$\% & $67.2$\% & $70.4$\% & $71.8$\%\\
			High             & $15.1$\% & $23.8$\% & $20.2$\% & $18.6$\%\\
			\botrule
		\end{tabular}
	\end{center}
\end{table}
To analyze the improvement in $SQI$ from the installation of an
additional fire station at the SW location, we show histograms of
$SQI$ values of all the properties considering only the existing fire
station and considering the existing station along with SW candidate,
as per~(\ref{eq:sqi}) (see Fig.~\ref{fig:sqi_hist_ex_sw}).  We see
that the addition of the SW candidate location increases the density
towards the head of the distribution, thus improving the number of
properties with low $SQI$ (better service quality).
Similarly, Fig.~\ref{fig:sqi_hist_sw_ne} and
\ref{fig:sqi_hist_sw_s} show the $SQI$ histogram comparison
between SW, NE and SW, S candidates. We observe that the improvement by
installing SW candidate, which is visible from the increased density
towards the head of the distribution, is clear from all the plots.

\begin{figure}[!h]
	\centering
	\captionsetup[subfigure]{labelformat=empty}
	\subfloat[]{\includegraphics[scale=0.28,trim={1.05cm 3.6cm 1.84cm 1.25cm},clip]{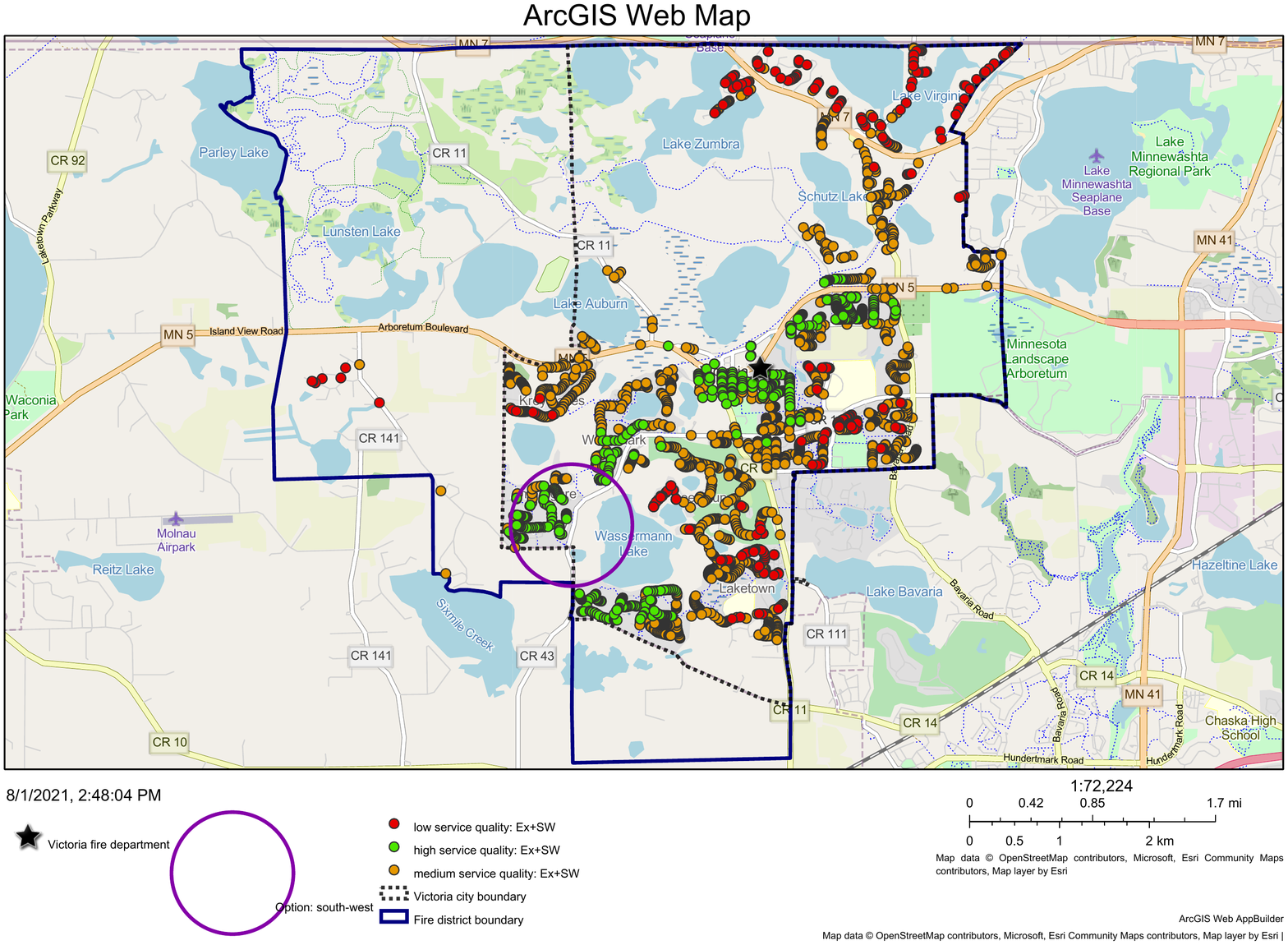}}\hspace{0cm}\vspace{-0.8cm}
	\subfloat[]{\includegraphics[scale=0.45,trim={1.05cm 0.2cm 16cm 18cm},clip]{map_service_quality_ex_sw.pdf}}
	\caption{Visual comparison with Fig.~\ref{fig:map_service_quality} reveals that installation of additional fire station at SW location would significantly improve the number of properties with high fire service quality (marked in green dots). Existing facility is marked with black star icon and the potential future fire station at SW location is shown with large circle to indicate broader zone. Blue solid line indicates fire district boundary. Black dashed line indicates Victoria city boundary}
	\label{fig:map_service_quality_ex_sw}
\end{figure}

\begin{figure}[!h]
	\centering
	\includegraphics[scale=0.9,trim={0.18cm 0cm 0.2cm 0.07cm},clip]{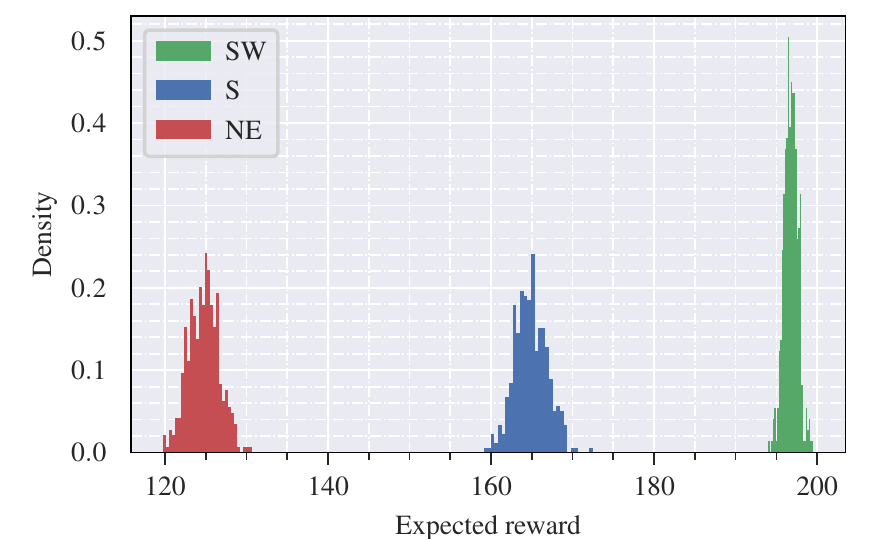}
	\caption{Histogram of expected reward accumulated by the candidates SW, S, and NE for 400 episodes each with 200 iterations. SW has the maximum expected reward. Variation of expected reward is significantly lesser in case of SW, compared to S and NE, as can be observed by narrow standard deviation of the histogram corresponding to SW. Thus, selection of SW location as the optimal choice to install a new fire station, as determined by our proposed algorithm has the highest confidence.}
	\label{fig:expected_reward}
\end{figure}

We conclude our analysis by illustrating the geographical distribution
of properties categorized, as per~(\ref{eq:sqi_category}), with both
the existing fire station and the additional fire station at SW location. The
plot is shown in Fig.~\ref{fig:map_service_quality_ex_sw}.
A visual comparison of Fig.~\ref{fig:map_service_quality} and
\ref{fig:map_service_quality_ex_sw} reveals that the installation of
the SW candidate significantly reduces the number of properties with
poor service quality. Note that the concentration of poor service
quality properties in the northeast and southern parts of the city is
difficult to avoid with the installation of only one additional fire
station due to longer travel times from either the existing facility
or from the SW candidate location. The longer travel time is mainly
attributed to the presence of lakes, terrain conditions, and the road
network.
\textcolor{black}{Next, we employ our proposed two-stage stochastic optimization model, as described in Section~\ref{subsec:stoch_opt_formulation}, to characterize the confidence in our decision to locate a new fire station at the SW location, which is apparent from our previous discussion. We run our proposed Algorithm~\ref{alg:opt_cand} in $400$ episodes each with a maximum iteration number $t_{max}=200$. The catchment areas and the candidate locations are same as described in the previous sections. The $\epsilon$ value is taken to be $0.7$. Fig.~\ref{fig:expected_reward} shows the density histogram plot of the expected rewards of the candidates SW, S, and NE for $400$ episodes. The histograms are plotted with a bin size of $40$. It can be observed that SW location accumulates highest expected reward followed by S and NE respectively. Further, the narrower histogram corresponding to SW location indicates lower standard deviation of the expected rewards, indicating a higher confidence in the decision to choose SW as the optimal location to install a new fire station.Therefore, we propose the SW location to VFD for the installment of an additional fire station.}

\section{Conclusions}\label{sec:conclusions}
In this work we present a systematic approach for locating new fire stations,
taking into account demand prediction, travel time, service
area coverage, and service redundancy. We propose a machine learning
model based on the random forest \textcolor{black}{and extreme gradient boosting} algorithms to predict service demand,
which can then be used to enhance the community safety. We present a detailed comparison between the performance of the models to select the better model for further study. \textcolor{black}{We observe that extreme gradient boosting model outperforms the random forest model}. With
the help of the machine learning model, we corroborate the influence
and the importance of the key factors governing fire service demand
associated with the properties. The model is further utilized to
define a service quality index to quantify quality of service, which
accounts for both demand prediction and travel time from fire stations
to demand locations. The quality index is utilized in a travel time
based DBSCAN algorithm to identify candidate locations for new fire
stations. Finally, we present a service quality index based optimization model that is derived from
the maximum coverage optimization problem, considering demand
prioritization and service area redundancy. We apply our proposed
methodology to select a location for a new fire station in the city of
Victoria, MN, USA. Further, we employ a two-stage stochastic optimization model to characterize the confidence in our decision outcome. We recommend that a future fire station be located
in the southwest portion of the city. This location would best
serve the community and improve the public safety standard.

\backmatter

\bmhead{Acknowledgments}

We would like to acknowledge Sarah Tschida, Program Coordinator for the Resilient Communities Project (RCP), University of Minnesota, Twin Cities, USA, for her assistance and efforts in maintaining an efficient collaboration between the Victoria Fire Department and University of Minnesota researchers. We would like to sincerely thank Victoria city council members for providing us with the opportunity to work on this project and their valuable feedback on the research. We would also like to thank GIS technicians at the Center for Urban and Regional Affairs (CURA), University of Minnesota, Twin Cities, for answering our queries about ArcGIS.

\section*{Declarations}
\begin{itemize}
\item Funding: No funds, grants, or other support was received.
\item Conflict of interest/Competing interests: The authors declare that they have no conflict of interest.
\item Availability of data and materials: Not applicable
\item Code availability: Not applicable
\item Ethics approval: Not applicable
\item Consent to participate: Not applicable
\item Consent for publication: Not applicable
\end{itemize}
\bibliographystyle{sn-basic}
\bibliography{references_springer}


\end{document}